\documentclass[10pt,twocolumn,letterpaper]{article}

\usepackage{iccv}      

\usepackage[dvipsnames]{xcolor}

\usepackage[pagebackref,breaklinks,colorlinks]{hyperref}
\usepackage{stfloats} 
\usepackage{multirow} 
\usepackage{algorithm}
\usepackage{algorithmicx}
\usepackage{algpseudocode}
\usepackage{stfloats}
\usepackage{graphicx}
\usepackage{xcolor}
\usepackage{comment}
\usepackage{amsmath}
\usepackage[bottom]{footmisc}
\usepackage{tabularx}
\usepackage{stfloats}
\usepackage{color}

\usepackage{tcolorbox}
\tcbuselibrary{breakable}
\definecolor{my_green}{RGB}{40,154,121}
\definecolor{my_red}{RGB}{176,46,46}
\definecolor{iccvblue}{rgb}{0.21,0.49,0.74}

\def\paperID{10683} 
\def\confName{ICCV}
\def\confYear{2025}

\title{LinkTo-Anime: A 2D Animation Optical Flow Dataset from 3D Model Rendering}

\author{
 Xiaoyi Feng$^{2*}$ \quad Kaifeng Zou$^{1*}$ \quad Caichun Cen$^{2}$ \quad Tao Huang$^{1}$ \quad Hui Guo$^{2}$\\
Zizhou Huang$^{1}$ \quad Yingli Zhao$^{4}$ \quad Mingqing Zhang$^3$ \quad Ziyuan Zheng$^{1}$ \\ DiweiWang$5$ \quad Yuntao Zou$^{2}$ \quad Dagang Li$^{2\dagger}$ \\
\small{$^1$ Link-To, Shenzhen, China \quad $^2$ Macau University of Science and Technology, Macau, China \quad} \\ \small{$^3$ The Chinese University of Hong Kong, HongKong, China \quad $4$ Wuzhou University, Guangxi, China} \\ \small{$5$ Université de Strasbourg, France} \\
\small{$^*$ Equal contribution \quad $^\dagger$ Corresponding author}
}

\begin{document}

\twocolumn[{
\renewcommand\twocolumn[1][]{#1}
\maketitle
\begin{center}
    \captionsetup{type=figure}
    \includegraphics[width=0.8\textwidth]{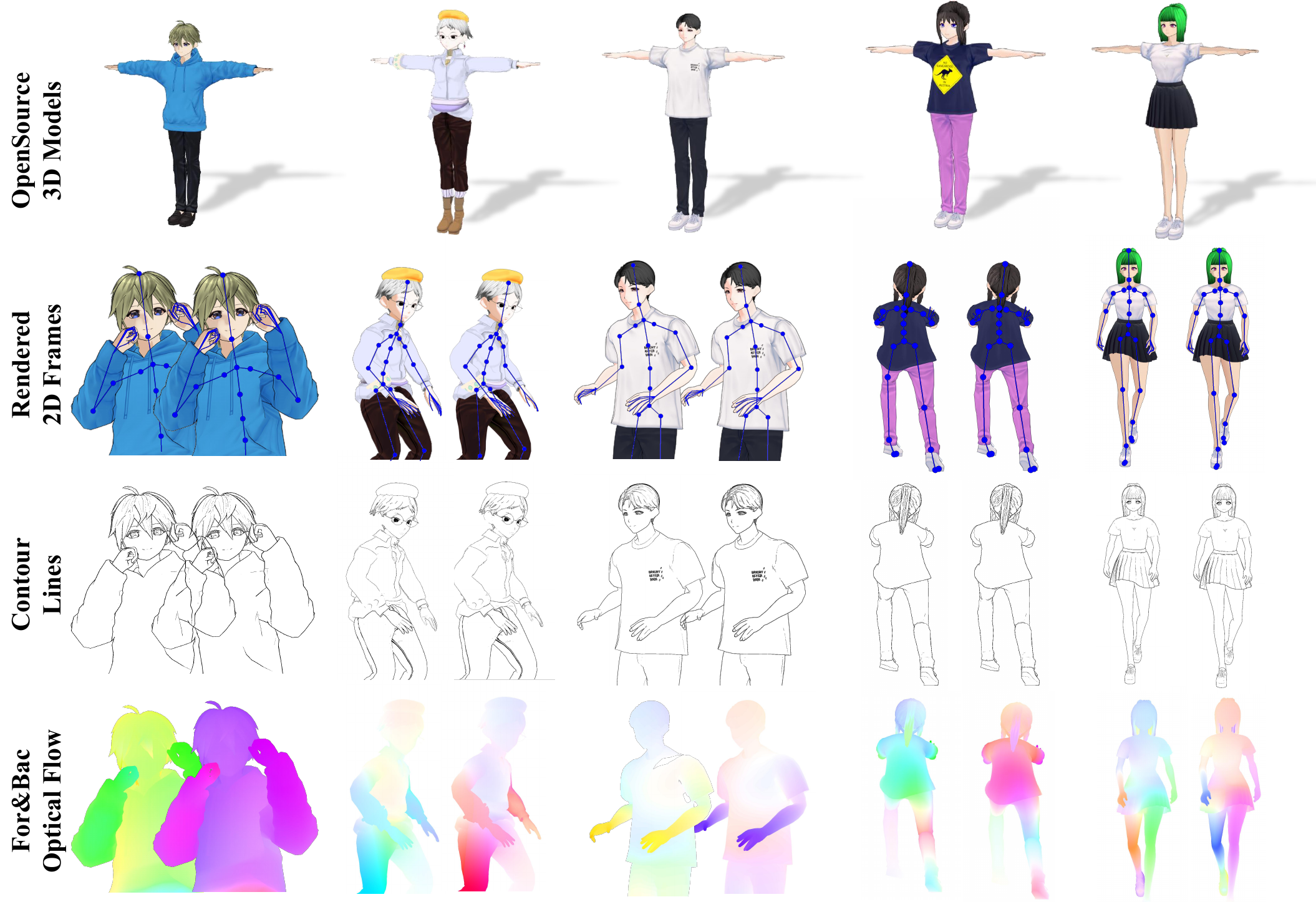}
    \vspace{-5pt}
    \captionof{figure}{We construct a dataset named \textbf{LinkTo-Anime} by augmenting open-source 3D models with manual model refinement, shading. We animate them using Mixamo skeletons to generate diverse character motions. The frames are rendered from multiple viewpoints, the resulting colored and line drawing resemble two styles in 2D cel animation production. Also, we provide pixel-level correspondence as the ground truth for optical flow estimation.}
    \vspace{10pt}
\end{center}
}]

\begin{abstract}
  Existing optical flow datasets focus primarily on real-world simulation or synthetic human motion, but few are tailored to Celluloid(cel) anime character motion: a domain with unique visual and motion characteristics. To bridge this gap and facilitate research in optical flow estimation and downstream tasks such as anime video generation and line drawing colorization, we introduce LinkTo-Anime, the first high-quality dataset specifically designed for cel anime character motion generated with 3D model rendering. LinkTo-Anime provides rich annotations including forward and backward optical flow, occlusion masks, and Mixamo Skeleton. The dataset comprises 395 video sequences, totally 24,230 training frames, 720 validation frames, and 4,320 test frames. Furthermore, a comprehensive benchmark is constructed with various optical flow estimation methods to  analyze the shortcomings and limitations across multiple datasets. Data are available at ~\url{https://huggingface.co/datasets/LecterF/LinkTo-Anime}.

\end{abstract}

\section{Introduction}
\label{intro}
\begin{figure*}[t]
    \centering
    \includegraphics[width=\textwidth]{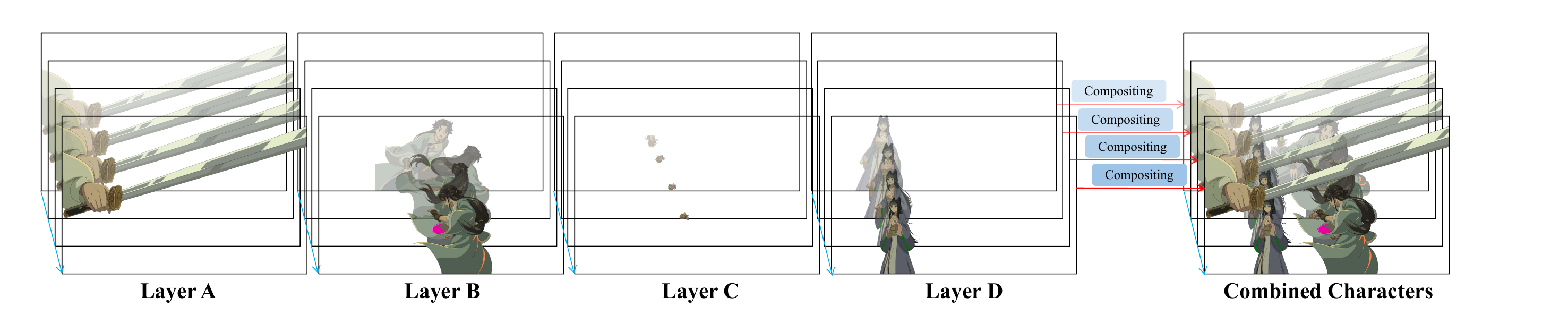}
    \caption{ Illustration of the compositing process in cel animation.
During production, different layers, such as characters and props are first processed independently for tasks like inbetweening and colorization. Once completed, all layers are composited to form the final scene. In this illustration, blue arrows indicate the motion trajectory from Frame 1 to Frame 4.
    }
    \label{fig:figurelayer}
    \centering
\end{figure*}

Celluloid (cel) animation, a type of traditional frame-by-frame animation, remains a cornerstone of the global animation industry, powering a vast range of productions from feature films to television and games.
In recent years, utilizing newly developed digital tools such as Toonz~\cite{toonz}, Adobe Flash~\cite{adobe}, and Clip Studio Paint~\cite{csp}, cel animation production has transitioned into the computer-assisted era, resulting in improved production efficiency~\cite{Cel_animation}.
To achieve efficient production, cel animation adheres to a well-organized process. Initially, scenes are drafted in layers; each character or dynamic object is depicted on an individual transparent sheet (cel), while the other areas usually remain blank or white. After completing each layer individually, the animator will check them and merge all layers together (composition), as shown in Fig.~\ref{fig:figurelayer}. This workflow gives the production files their unique characteristics, blank backgrounds. In addition, cel animation production files have another distinctive characteristic. The camera configurations generally remain stationary within each cut, a small video unit in cel animation production~\cite{cut}, with movement expressed entirely via character actions rather than camera motion, as demonstrated in Fig.~\ref{fig:figurelayer}. This practice enables animators to reutilize the background in each cut, thereby optimizing efficiency, given that backgrounds are more challenging to illustrate. 
Under such conditions, the visual characteristics of images in cel animation production differ significantly from those in real-world videos and final produced anime content.

\begin{figure}[t]
    \centering
    \includegraphics[width=\linewidth]{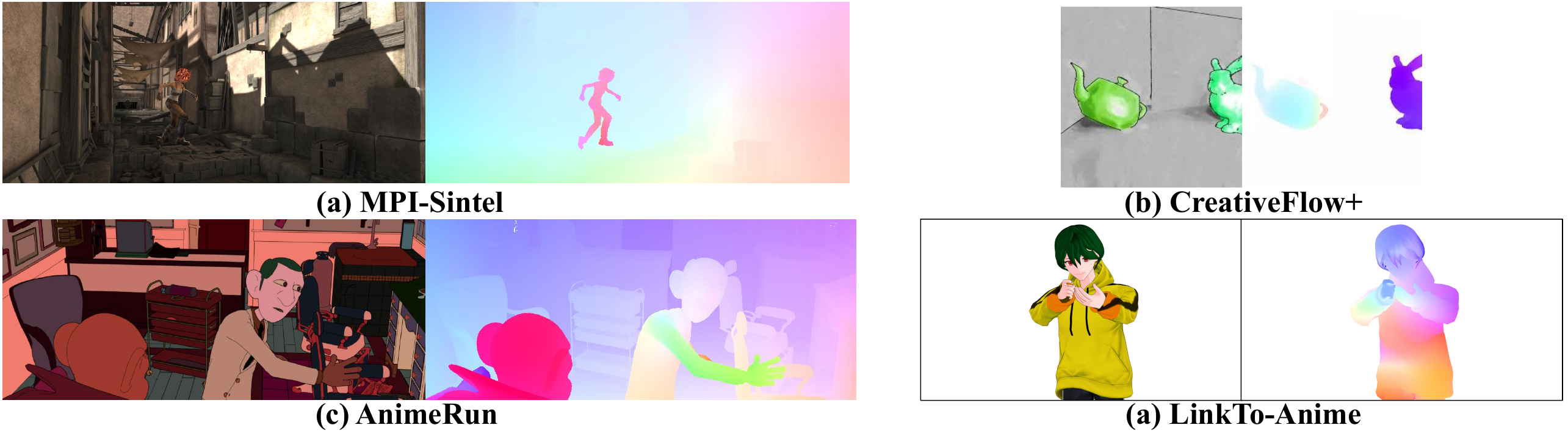}
    \caption{ Comparative visualization of images and optical flow fields from four datasets: (a) MPI-Sintel~\cite{sintel}, (b) CreativeFlow+~\cite{creative}, (c) AnimeRun~\cite{animerun}, and (d) LinkTo-Anime (Ours). 
    }
    \label{fig:figuredatasets}
    \centering
\end{figure}

With rapid advances in deep learning, many recent works~\cite{ant,pbcarticle,pbcconference,lvcd,manganinja,line_inbetween,tooncrafter} have proposed learning-based systems to automate parts of the cel animation production, including tasks such as in-between (frame interpolation)~\cite{line_inbetween,tooncrafter} and line drawing colorization~\cite{ant,pbcarticle,pbcconference}. Among the various techniques, optical flow prediction plays a critical role in both the training and inference stages. However, achieving high-quality predictions requires domain-specific datasets specifically created for cel animation production images. Although several datasets~\cite{creative,animerun} have been released for optical flow prediction in 2D animation, there remain notable discrepancies between the characteristics of these datasets and those images in the cel animation production. As illustrated in Fig.~\ref{fig:figuredatasets}, CreativeFlow+ dataset~\cite{creative} provides a large number of sequences with relatively clean backgrounds that resemble animation. However, it lacks high-resolution and semantically rich character appearances, which are essential in real-world anime production. On the other hand, the AnimeRun data set~\cite{animerun} improves over general-purpose data sets by applying cartoon-specific optimizations such as flattening and edge extraction. Yet, it suffers from a small scale and background movement that do not reflect actual cel animation characteristics.
Facing these challenges, we introduce \textbf{LinkTo-Anime}, the first high-quality optical flow dataset tailored specifically to cel animation character movements. In our dataset, for each frame, we provide both the rendered colored frame and the processed line-art version. The specific rendering procedure for generating line art is detailed in Sec.~\ref{renderoption}.
For the labels, we include both forward and backward optical flow computed from Blender's vector motion, and occlusion masks are provided.
Each cut in our dataset is generated by animating a 3D character model using motion driven by a Mixamo skeleton and rendering the images frame by frame. The 3D character models are sourced from publicly available assets on the VRoid hub~\footnote{\href{https://vroid.com/en}{https://vroid.com/en}}.
To ensure diversity in both appearance and motion, we selected a wide range of 3D models and rendered each of them from many different camera views: upper-body shot, full-body shot, lower-body shot, front view, back view,~\etc. Moreover, to better match the data distribution seen in cel animation production, we carefully constrained the rendering setup such that backgrounds remain uniformly white, which motion occurs only in the foreground regions.
Compared to previous datasets, our dataset offers the following contributions:
1. Images align with cel animation production: our dataset exhibits data characteristics that are closer to cel animation production images, such as white backgrounds and rendered images that resemble the style of cel animation.
2. Large-scale labeled images with high resolution and fine-grained character details: Our dataset features high-resolution images and preserves fine character details.
3. High diversity through multiview character renderings: Our data set provides multiple rendered views per character, allowing greater diversity in appearance and pose.
We split our dataset into three parts: train, validation, and test. For evaluation, we perform optical flow benchmarking on the test set of our dataset. In addition, we construct a private anime industry dataset named \textbf{Cel}, with 1100 images collected from various anime studios to validate the effectiveness of the proposed data set. We train models using a variety of optical flow datasets and prediction methods. Experimental results demonstrate that our method not only improves optical flow estimation on images from this domain, but also enhances the performance of downstream models.

In summary, our work makes the following contributions:
1. We construct an optical flow dataset specifically designed for cel animation production, providing accurate flow label annotations.
2. We conduct comprehensive benchmarking across multiple datasets to demonstrate the feasibility of our dataset, and further validate its effectiveness on cel animation production applications.

\begin{table*}[t]
\centering
\caption{Optical flow statistics across different datasets. We report the proportion of valid flow (i.e., pixels with flow magnitude \textgreater 0.01). Ratio s \textless 10 indicates the percentage of valid flow pixels with a motion speed less than 10 pixels. Ratio 10–50 refers to the percentage of pixels with motion speed between 10 and 50 pixels, and Ratio \textgreater 50 represents those with motion speed greater than 50 pixels.}
\label{tab:table_datasets}
\resizebox{\textwidth}{!}{%
\begin{tabular}{lccccccc}
\hline
Dataset        & Modality & Clips & Frames & Resolution        & Ratio s \textless10 & Ratio s10-50 & Ratio s\textgreater50 \\ \hline
MPI-Sintel~\cite{sintel} & Video    & 35  & 1628   & 436$\times$1024         & 68.35\%        & 24.2\%            & 7.45\%          \\
CreativeFlow+~\cite{creative} & Video    & 3000  & 134k   & 1500$\times$1500         & -           & -            & -          \\
AnimeRun~\cite{animerun}       & Video    & 30    & 3k  & 429$\times$1024, 540$\times$960 & 64.12\%      & 27.62\%       & 8.11\%      \\
LinkTo-Anime     & Video    & 395   & 30K & 1440$\times$2560         & 58.28\%      & 36.25\%       & 4.36\%     \\ \hline
\end{tabular}
} 

\end{table*}

\begin{figure*}[t]
    \centering
    \includegraphics[width=\textwidth]{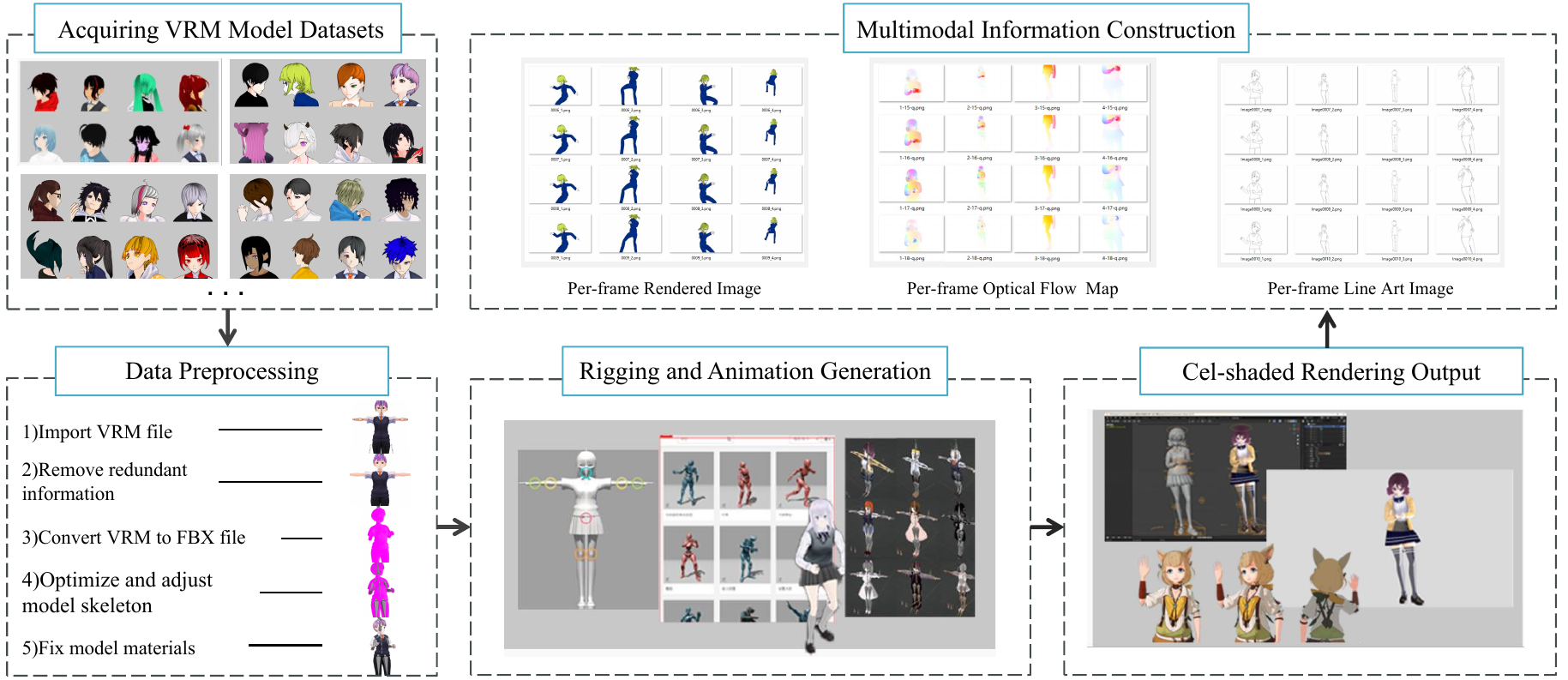}
    \caption{\textbf{Overview of our data preparation and rendering pipeline.} The process consists of five stages: (1) Acquiring 3D models from open-source platforms (e.g., Vroid); (2) Preprocessing models to clean and restructure them; (3) Rigging and generating motions using Mixamo; (4) Aadapting models to Cel-Animation style by creating line art and applying shading; (5) Exporting multiple outputs including colored frames, line drawings, and optical flow annotations.}

    \label{fig:figurerenderprocess}
    \centering
\end{figure*}

\section{Related Work}
\label{gen_inst}
\subsection{Celluloid (Cel) Animation}
Celluloid animation, emerging as a form of traditional animation in the 1920s, has become a crucial foundation for the modern animation industry.~\cite{Cel_animation}. In this field, animators consistently adhere to a series of systematic steps to produce cartoons, encompassing scripting, storyboarding, layout, in-between, colorization, composition, cutting, ~\etc~\cite{Cel_animation}. Many studies have been introduced to provide automated or supportive solutions for anime production~\cite{ant,pbcarticle,pbcconference,lvcd,anime_director,manganinja,loftsdottir2022sketchbetween,siyao2023deep,line_inbetween,wang2023coloring,tooncrafter,lvming_scribble,colorflow}. For example, the Anime-director~\cite{anime_director} focuses on the sequential creation of coherent story scripts and the semantics of images, contributing to automated storyboarding. On the other hand, tooncrafter~\cite{tooncrafter} employs diffusion models to interpolate intermediate frames from given keyframes for in-between. Furthermore, techniques such as BasicPBC~\cite{pbcconference, pbcarticle}, LVCD~\cite{lvcd} and MangaNinja~\cite{manganinja} have been developed to facilitate automated colorization of black-and-white line drawings, thus enhancing the effectiveness of colorization.

\subsection{Optical Flow}
In recent years, significant interest has been generated in the field of optical flow, which supplies motion data at the pixel level, prompting the creation of various techniques~\cite{dong2024memflow,huang2022flowformer,jeong2023distractflow,gma,shi2023flowformerpp,sun2018pwc,sun2022disentangling,raft,xu2022gmflow,zhao2020maskflownet, zhu2025cogcartoon} and numerous datasets~\cite{sintel,gaidon2016virtual,flyingthings,creative,animerun}. Optical flow reflects pixel-level correspondences, thus many researchers opt for computer-rendered synthesis methods to generate such data. In~\cite{sintel}, Buter~\etal, created a dataset utilizing 3D video sequences with rendered optical flow data. Subsequent research led to the creation of large-scale datasets such as FlyingThings3D~\cite{flyingthings}, Virtual KITTI~\cite{gaidon2016virtual}, and others for extensive training. Within the realm of cel animation, optical flow can enhance numerous downstream processes, such as direct frame interpolation~\cite{upr,narita2019optical,niklaus2020softmax,rife,xu2019quadratic} or guiding this interpolation~\cite{flowi2v}. Moreover, it aids colorization methods as in~\cite{pbcarticle,pbcconference}. The two datasets most similar to the cel animation attribute are CreativeFlow+~\cite{creative} and AnimeRun~\cite{animerun}. CreativeFlow+~\cite{creative} employs different line and shading styles to render 3D models. However, this dataset is limited by its overly simplistic 3D models, which lack intricate details, and rendered images that do not closely resemble cel animation. Alternatively, Siyao~\etal rendered 3D cartoons in 2D to compile a 2D animation dataset called AnimeRun~\cite{animerun}, using FLAT rendering to convert 3D data into a cel animation format. Still, this dataset has its limitations, as the FLAT rendering induces detail loss and all images are full-scene frames. As mentioned in Sec.~\ref{intro}, images used in cel anime production, their background always remain blank. Consequently, the style of AnimeRun~\cite{animerun} differs from the inherent image style of cel anime production. In this context, we intend to develop a dataset with visual styles that match the attributes of images used in cel anime production to facilitate tasks like coloring and interpolation. Consequently, this paper introduces LinkTo-Anime, a high-quality optical flow dataset created by rendering intricate 3D models specifically designed for the cel anime style.
\section{LinkTo-Anime Dataset}
\label{Linto30kSec}
Optical flow, as a pixel-level correspondence label, is not only difficult to obtain in real-world scenarios, but also infeasible to acquire from standard 2D animation. As a result, creating such a dataset necessarily relies on rendering 2D images from 3D models with motion information. In this section, we detail our exploration and choices regarding rendering strategies and line-art synthesis. We also provide a comparative analysis between our dataset, two existing anime-style optical flow datasets, and a private dataset collected from actual animation production in Sec.~\ref{data_analysis}.
\subsection{Rendering Options}
\label{renderoption}

\begin{figure}[t]
    \centering
    \includegraphics[width=\linewidth]{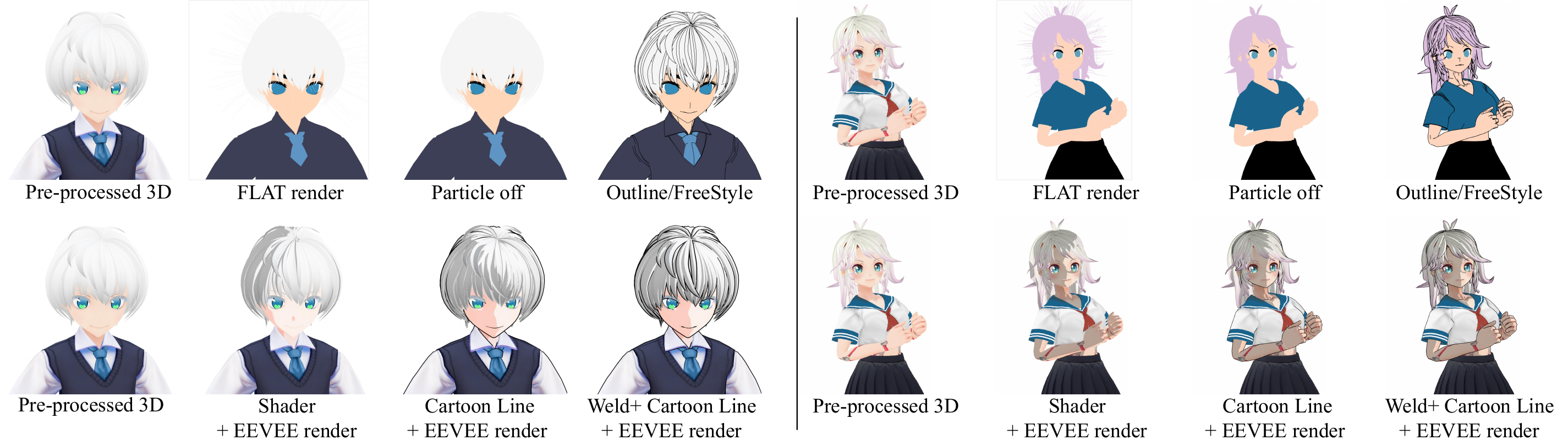}
    \caption{Visualization of two rendering pipelines. Each row represents a complete rendering process, and each column shows an intermediate step. Two models are shown side by side for comparison. The first pipeline uses FLAT render with Freestyle lines after removing particles, while the second applies shading, then generate contours with Cartoon Line Creator, followed by line refinement using a weld modifier. The second pipeline preserves much more detail and is adopted in our final dataset.}

    \label{fig:figurerenderexp}
    \centering
\end{figure}
\textbf{Rendering Styles}
Before rendering and processing 3D models, we first analyzed several key characteristics of cel animation. As mentioned earlier, during the production process, regions outside the main subject are typically left blanked. Secondly, 2D animations are usually composed of unique color regions enclosed by clean line art~\cite{pbcarticle}. These color regions often exhibit flat value. Thirdly, with the advancement of production techniques, modern cel animation features characters with finer details and more refined appearances. Therefore, when constructing our dataset, our primary goal was to render images that not only resemble the color and line style of cel animation, but also preserve as much structural and visual detail as possible.
Thus, we explored several combinations of rendering techniques. Before that, the very frist step is the pre-process of downloaded open-source 3D models. Open-source 3D models often come with various issues, such as non-3D hair represented by planar strips, incorrect skeleton assignments, and other structural problems that can significantly affect the quality of render results. Therefore, after downloading these models, we manually applied pre-processing steps, such as removing non-3D hair structures and replacing them with fully 3D representations, to ensure better rendering quality. After that, as shown in Fig.~\ref{fig:figurerenderexp}, we experimented with the combination of ``FLAT'' rendering and ``FreeStyle'' line rendering as mentioned in \textbf{AnimeRun}~\cite{animerun}. We found that although this approach produced uniform color regions, it caused significant loss of visual details, such as the interior features of the eyes or the fine structures around the collars in Fig.~\ref{fig:figurerenderexp}. Consequently, we decided not to adopt the ``FLAT'' rendering approach. Instead, we choose to use the rendering method we explore as shown in the second row of Fig.~\ref{fig:figurerenderexp} to balance the style consistency with cel animation and detail preservation.

\textbf{Line Generation}. To generate contour lines that are as continuous and stable as possible without relying on the ``Flat'' Rendering, we experimented with several approaches. Including Blender's built-in FreeStyle rendering, the use of the Materialization Modifier to simulate line effects, and the third-party plugin Cartoon Line Creator. After a comparative analysis, we ultimately adopted the Cartoon Line Creator approach, as it produced the most uniform and stable lines overall.
To further enhance the continuity and visual quality of the lines, we applied the Welding Modifier prior to the libe art output. This operation automatically merges nearby vertices, effectively correcting discontinuities caused by modeling imperfections or line generation artifacts, thereby improving the visual consistency of the rendered lines.

\textbf{Shading}. To mimic the shadows in cel animation, we incorporated shading during the rendering. Lambertian diffuse shading was employed to generate shadow patches, a normal map was added to represent surface details and serve as a shading mask, thereby enhancing light-dark contrast.

\textbf{Backbone and Motion}. 
To animate our models, we ensured that each one was properly rigged with skeletons. For models that already included Mixamo skeletons, we inspected and fixed any issues in the bone structure. For those without any skeletal rigging, we used Mixamo~\footnote{\href{https://www.mixamo.com/}{https://www.mixamo.com/}} platform to automatically rig skeletons. We also used the motion sequences provided by Mixamo to animate the models. However, we found that automatic skeleton rigging could not produce reasonable motion in fine structures such as hair and clothing. To address this problem, we manually added supplementary bones into the hair and adjusted them to mimic motion, such as the swaying of hair.

\begin{figure}[t]
    \centering
    \begin{subfigure}[t]{0.24\linewidth}
        \centering
        \includegraphics[width=\linewidth]{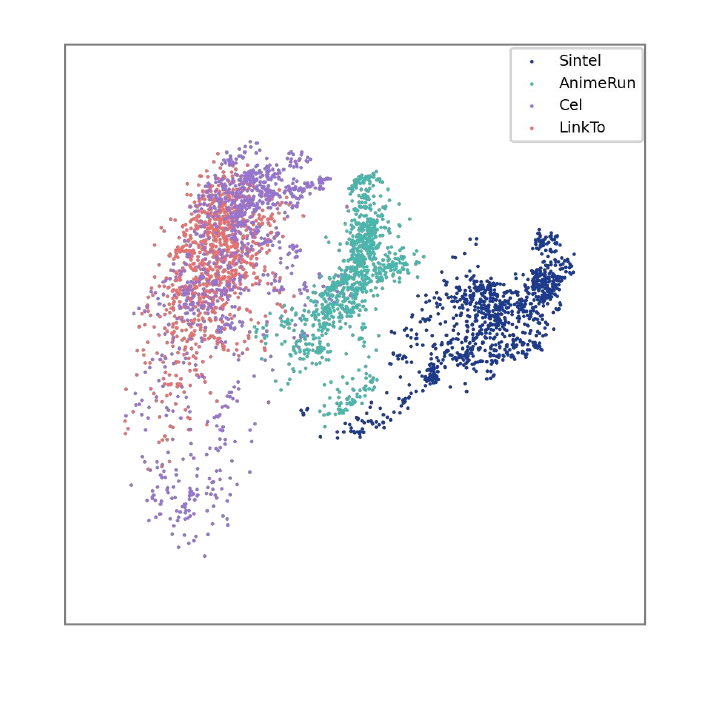}
        \caption{PCA.}
        \label{fig:sub-pca}
    \end{subfigure}
    \hfill
    \begin{subfigure}[t]{0.24\linewidth}
        \centering
        \includegraphics[width=\linewidth]{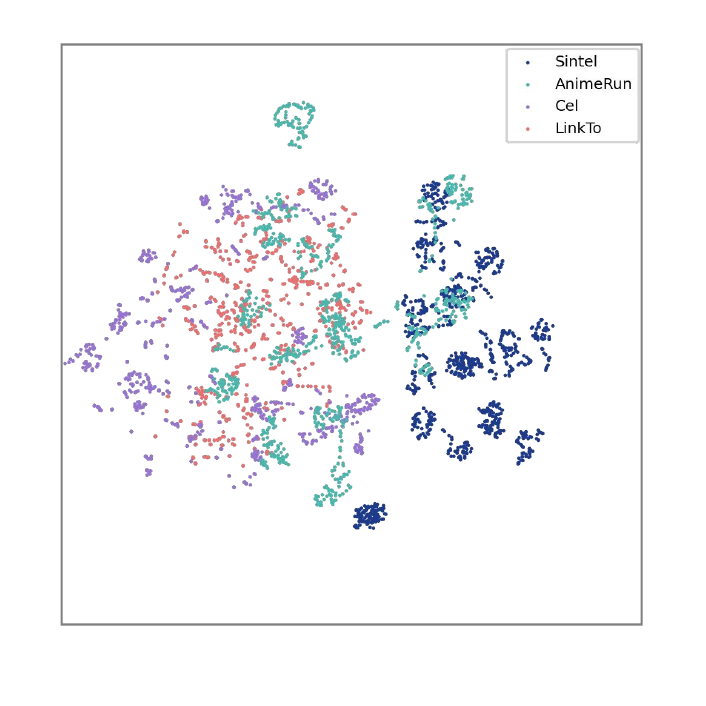}
        \caption{t-SNE.}
        \label{fig:sub-tsne}
    \end{subfigure}
    \begin{subfigure}[t]{0.24\linewidth}
        \centering
        \includegraphics[width=\linewidth]{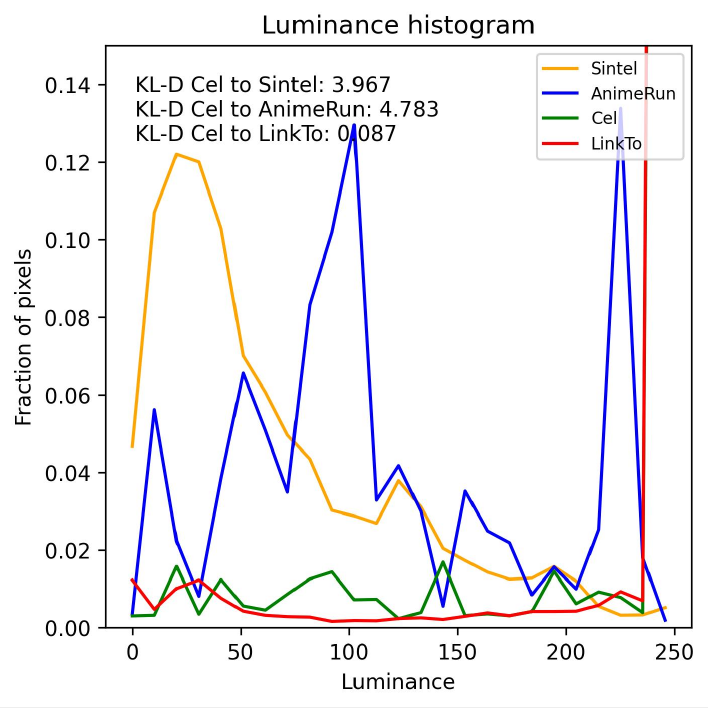}
        \caption{Luminance}
        \label{fig:sub-lumi}
    \end{subfigure}
    \begin{subfigure}[t]{0.24\linewidth}
        \centering
        \includegraphics[width=\linewidth]{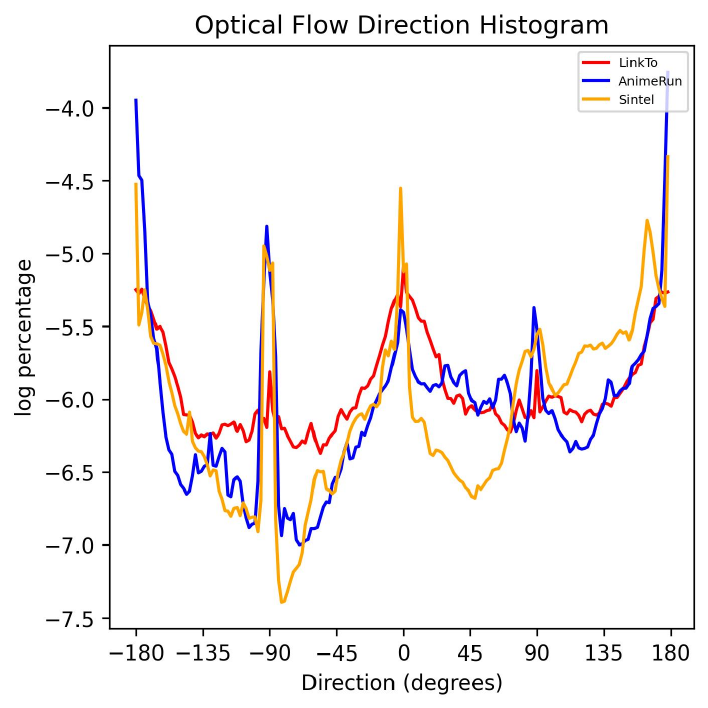}
        \caption{Direction}
        \label{fig:sub-direct}
    \end{subfigure}
    \caption{Comparison on dataset distributions. (a) PCA: LinkTo-Anime and Cel significantly overlap in their data distribution. (b) t-SNE: The distributions of the LinkTo-Anime and the Cel largely overlap. (c) Luminance histogram:Among~\textcolor{orange}{Sintel},~\textcolor{blue}{AnimeRun},~\textcolor{red}{LinkTo-Anime},and~\textcolor{green}{Cel}. the histogram of ~\textcolor{red}{LinkTo-Anime} is closest to that of the anime-indsutry dataset,~\textcolor{green}{Cel}.}
    \label{fig:fig:distribute_compare}
\end{figure}

\textbf{Viewpoints and Camera Setting}. During the rendering process, we adopted various camera angles and focal lengths. The focal lengths include Medium Long Shot (MLS), Medium Shot, and Medium Close-Up,~\etc. The camera angles cover Full Body Shot, Upper Body Shot, Lower Body Shot, and Back View Shot,~\etc. For each video, we mimicked individual cuts as commonly seen in animation production, where the camera position is fixed for each cut and the focal length remains unchanged.



\subsection{Video Information and Data Split}
As shown in Tab.~\ref{tab:table_datasets}, our dataset consists of 29270 frames derived from 395 video clips. These contents were rendered using 80 different 3D models from the vroid hub~\footnote{\href{https://vroid.com/en}{https://vroid.com/en}} following Panic3D~\cite{chen2023panic3d}. For each model, we rig a Mixamo skeleton and assign a specific motion, such as boxing, dancing,~\etc. Each model is rendered with five different viewpoints, resulting in five clips, with each clip containing no less than 72 frames.
As to the data split, the dataset is divided into three subsets: training, validation, and testing, containing images rendered from 65, 2, and 13 3D models, respectively. To maintain consistency, all data are rendered at a fixed resolution of 1440×2560.

 \subsection{Data Analysis and Comparison}
 \label{data_analysis}
To analyze the data distribution of our method, we constructed a private dataset composed entirely of production files from cel animation processes called \textbf{Cel}, collected from various anime studios. This dataset contains 1,100 frames.
We compared the distribution of four datasets: Sintel, AnimeRun, our private anime-industry dataset (Cel), and the proposed LinkTo-Anime. To evaluate their distributions, we employed two different dimensionality reduction techniques for visualization. As shown in Fig.~\ref{fig:sub-pca}, we first visualize differences in data distribution using Principal Component Analysis (PCA)~\cite{pca}. Our dataset overlaps significantly with the distribution of the anime-industry data, while AnimeRun lies relatively close to them. In contrast, Sintel is noticeably distant from the others.

In the same way, as illustrated in Fig.~\ref{fig:sub-tsne}, the visualization of t-SNE further supports this observation~\cite{tsne}. The arrangement of our dataset is very similar to that of the Cel dataset. In contrast, AnimeRun is more separated, and Sintel is the most outlying. This result suggests that our dataset is most similar to actual anime-industry production data, with Sintel being the least similar, probably due to its design as a naturalistic animation, which involved additional data augmentations.

We also performed a luminance analysis. As shown in Fig.~\ref{fig:sub-lumi}, except for the white background, the pixel values in our dataset are relatively evenly distributed throughout the range 0 to 254. Furthermore, we calculated the KL-Divergence between the luminance histograms of the Cel dataset and the other three datasets: Sintel (3.960), AnimeRun (4.853), and LinkTo-Anime (0.082). These results demonstrate that the luminance distribution of our dataset is closest to that of the Cel dataset. 
Moreover, we conducted an optical flow direction analysis across three datasets in Fig.~\ref{fig:sub-direct}. The flow directions in the compared datasets are mostly concentrated around 0, 90, -90, and 180 degree;, however, our dataset exhibits a more uniformly distributed curve, indicating greater diversity in motion directions.

\begin{table*}[t]
\centering

\caption{\textbf{Quantitative outcomes of optical flow fine-tuned on various datasets with different models}. `occ': occluded regions; `s\textless 10': region with speed below 10 pixels; `s10-50': region with speed ranging from 10 to 50 pixels; `s\textgreater50': region with speed exceeding 50 pixels. Datasets: S: Sintel\cite{sintel}; T: FlyingThing3D~\cite{flyingthings}; Cr: CreativeFlow+~\cite{creative}; R: AnimeRun~\cite{animerun}; LT: LinkTo-Anime.}
\resizebox{\linewidth}{!}{%
\begin{tabular}{l|c|cc|ccc|cc}
\hline
                        & \multicolumn{6}{c}{Our 3D Rendered Dataset Testset}  \\ \hline
Method                  & EPE(pix.) & occ.  & non-occ. & s\textless10  & s10-50 & s\textgreater50 & bg & fg \\ \hline
PWC            & 119.04    & 20.85 & 111.81  & 116.49 & 5.14   & 12.95  & 111.92 & 4.55 \\
PWC ft. T+S    & 8.47     & 18.57 & 8.38  & 8.64  & 4.24   & 12.01 & 8.50 & 3.63  \\
PWC ft. T+Cr   &  0.52     & 6.82  & 0.48  &  0.30  &  4.28 & 9.98 & 0.50 & 4.50 \\
PWC ft. T+R    & 15.23     & 18.74 & 15.21   & 15.83  & 3.14   & 6.96 & 15.33 & 3.31 \\
PWC ft. T+LT & 0.30      & 5.38  & 0.26  & 0.15  & 2.80   & 5.65 & 0.27 & 3.57 \\ \hline
RAFT               & 8.48      & 18.49 & 8.40    & 8.61  & 4.88  & 15.68 & 8.52 & 3.82 \\
RAFT ft. T+S       & 9.91     & 20.05 & 9.83  & 10.13  & 4.68  & 13.88 & 9.96 & 3.81 \\
RAFT ft. T+Cr      & -         & -     &  -       &  -      &  -      &  -  & -  & - \\
RAFT ft. T+R       & 48.62   & 20.67 & 48.84  & 50.97 & 2.57 & 5.49 & 48.97 & 3.21\\
RAFT ft. T+LT    & 0.20      & 3.81  & 0.17  & 0.09 & 2.13  & 4.33 & 0.18 & 2.68 \\ \hline
GMA                & 7.89    & 19.03 & 7.81  & 8.05  & 4.11  & 11.19 & 7.93 & 3.44\\
GMA ft. T+S        & 9.64    & 19.18 & 9.57 & 9.90  & 3.97  & 10.47 & 9.69 & 3.38 \\
GMA ft. T+Cr       &  1.89   & 8.01 & 1.84  & 1.81   & 2.87   & 7.28 & 1.87 & 3.76  \\
GMA ft. T+R        & 10.39  & 16.63 & 10.34  & 10.76  & 2.84  & 6.12 & 10.45 & 2.99 \\
GMA ft. T+LT     & \textbf{0.20 } & \textbf{3.78 } & \textbf{0.17 } & \textbf{0.08}  & \textbf{2.12}  & \textbf{4.32 } &\textbf{ 0.18} & \textbf{2.64} \\ \hline
GMFlow             & 6.47 & 17.60 & 6.38 & 6.51 & 4.45  & 14.70 & 6.49 & 3.73 \\
GMFlow ft. T+S     & 7.24 & 15.48 & 7.18 & 7.27 & 5.20 & 19.57 & 7.27 & 4.11 \\
GMFlow ft. T+Cr    & 0.59 & 6.42  &  0.55   & 0.27  & 5.71   & 14.99  & 0.56 & 4.75 \\
GMFlow ft. T+R     & 12.35 & 20.30 & 12.29 & 12.79 & 3.29 & 8.34 & 12.43 & 3.16 \\
GMFlow ft. T+LT  & 0.31 & 5.14 & 0.27 & 0.16  & 2.83 & 5.48 & 0.29 & 2.95 \\ \hline
\end{tabular}
}
\label{table1}
\end{table*}

\section{Experiments}
\label{experiment}
\subsection{Experimental Setting}
We conducted benchmark on the LinkTo-Anime test set. We selected four widely recognized optical flow prediction models. PWC-Net~\cite{sun2018pwc}, RAFT~\cite{raft}, GMA~\cite{gma}, and GMFlow~\cite{xu2022gmflow}. To evaluate generalization and adaptation, we fine-tuned each model on four datasets: Sintel\cite{sintel}, CreativeFlow+\cite{creative}, and AnimeRun\cite{animerun}, LinkTo-Anime. For brevity, we denote each setting as model ft. T+X, where T represents the base training set (mixing with 1:10 FLyingthings3D) and X indicates the fine-tuning dataset (\eg, ft. T+S for fine-tuning on Sintel, ft. T+Cr for CreativeFlow+, ft. T+R for AnimeRun and ft. T+LT LinkTo-Anime).
We provide both quantitative and qualitative evaluation results on the benchmark of LinkTo-Anime test set. For the performance on the private industrial dataset (\textbf{Cel}), where ground-truth optical flow is not available, we present only qualitative visualizations of the predicted results.
\begin{figure*}[t]
    \centering
    \includegraphics[width=\textwidth]{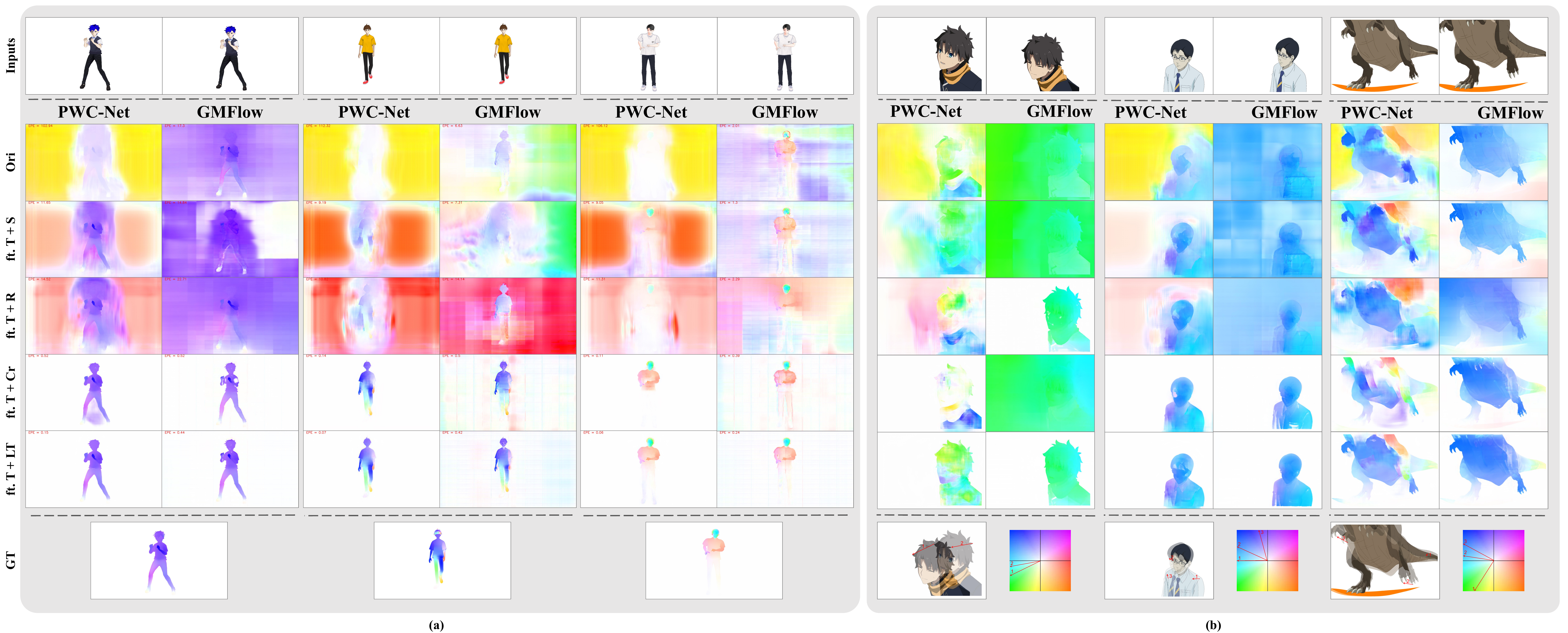}
\caption{\textbf{Visualization of optical flow predictions on the LinkTo-Anime test set (a) and Cel-dataset (b).} In (a), the model fine-tuned on LinkTo-Anime demonstrates better performance compared to those fine-tuned on other datasets such as Sintel~\cite{sintel}, CreativeFlow+~\cite{creative}, and AnimeRun~\cite{animerun}. It not only effectively reduces errors in the background regions but also improves prediction accuracy in fine-detailed areas. For (b), in the absence of ground truth, we manually annotated several sparse pixel movements (~\textcolor{red}{Arrows}) on image pairs. The correspondences between pixels in two frames were found based on the endpoints of black lines. For each pixel pair, we illustrated the motion direction on the optical flow color wheel. This serves as a coarse ground truth to roughly assess the accuracy of optical flow predictions.}
    \vspace{-10pt}
    \label{fig:figuremainplot1}
    \centering

\end{figure*}

\textbf{Settings.} 
For each optical flow method, we followed the training protocol of~\cite{animerun}, fine-tuning models for 40,000 iterations on the training sets of each dataset. Following standard practice, 10\% of FlyingThings3D~\cite{flyingthings} was included in every training run. The input resolution was 368×768, with a batch size of 16, a learning rate of $10^{-4}$, and a weight decay of $10^{-5}$. These settings were kept consistent across datasets. Due to the unavailability of the download link for CreativeFlow+\footnote{\href{https://www.cs.toronto.edu/creativeflow/}{https://www.cs.toronto.edu/creativeflow/}}, we used the pretrained weights provided in AnimeRun\footnote{\href{https://drive.google.com/drive/folders/16fA0w1vaaU4gD7QVKRSfTRUBejan0ib8}{https://drive.google.com/drive/folders/}}. Since the fine-tuned RAFT weights on CreativeFlow+ were not released,  we did not report results for this setting.
For evaluation metrics, we mainly used the end-point error (EPE), which computes the Euclidean distance between the predicted flow and the ground truth, serving as a direct measure of prediction accuracy. Occlusion occurs in many frames of our dataset due to body motion (\eg arms occluding the chest). To assess prediction quality in such regions, we calculated EPE separately for occluded and non-occluded areas, denoted as ``occ.'' and ``non-occ.'', respectively. Furthermore, to analyze performance across different motion magnitudes, we divided the flow pixels into three velocity ranges (\textless 10, 10–50, \textgreater 50) and foreground~\& background (``fg''~\& ``bg''). All experiments were conducted on NVIDIA RTX L20 GPU.


\subsection{Experimental Results and Analysis}

\textbf{Model ft. T+LT can improve the performance.}
In Tab.~\ref{table1}, the results of the four methods demonstrate that fine-tuning on our dataset consistently improves the optical flow prediction performance for all EPE metrics. For the overall EPE, GMA ft. T + LT  outperformed the second best (PWC-Net ft. T + Cr) by 0.10. RAFT ft. T + LT outperforming its original model by 8.28. PWC-Net ft. T + LT improved by 0.22 compared to its second-best (PWC-Net ft. T + Cr), and GMFlow ft. T + LT surpassed its corresponding second-best (GMFlow ft. T + Cr) by 0.28 points. The visualization results in Fig.~\ref{fig:figuremainplot1} also demonstrate the superiority of our dataset, as our predictions exhibit the cleanest backgrounds and the most accurate foregrounds.

\textbf{Optical flow estimation performance analysis.} 
We observed that models ft. T + Cr achieved an overall EPE similar to that fine-tuned on our dataset. This is mainly due to the comparable prediction of model ft. T + Cr and model ft. T + LT in the region of $s > 10$ where the white background is predominant. And this result can be attributed to the presence of pure backgrounds in both our dataset and CreativeFlow+\cite{creative}, which provides the models with more relevant and diverse training data to better handle such scenarios. In contrast, complex backgrounds present in the Sintel\cite{sintel} and AnimeRun\cite{animerun} datasets, models that are fine-tuned using these datasets tend to exhibit poor performance when it comes to estimating pure or white backgrounds. And this can lead to low overall EPE performance. The visualization results in Fig.~\ref{fig:figuremainplot1} also confirm this observation. 

However, when focusing on areas of higher motion (that is, $s \in [10, 50]$ and $s > 50$), the gap between model ft. T + Cr and model ft. T + LT becomes more apparent. For example, on PWC-Net, in these areas for the model fine-tuned on Our dataset is greater than model fine-tuned on T + Cr by 1.48 and 4.33, respectively. Comparing to our dataset, the use of simpler 3D models in CreativeFlow+~\cite{creative} may contribute to this outcome.

In some situation, T + R fine-tuning tends to degrade performance in regions of background area. Specifically, the fine-tuned RAFT, GMA, and GMFlow model on AnimeRun ranked last among all variants. his is likely because the background area of AnimeRun is not pure white, which is not consist to the characteristic of cel animation production images. 

\textbf{Analysis of Cel dataset results.}
 We evaluate the performance of models fine-tuned under different settings on our constructed anime-industry dataset, \textbf{Cel}. 
Since obtaining full ground truth annotations for the cel dataset is challenging, we manually annotated a few key points as coarse ground truth and primarily relied on visual comparisons to assess the performance of different fine-tuning strategies.
Fig.~\ref{fig:figuremainplot1} (b) illustrates the distinct benefits of our dataset. Firstly, compared to models pre-trained and fine-tuned on Sintel and AnimeRun, our dataset significantly enhances model performance in background areas and fine detailed areas compared to other datasets. For instance, in the second column, where the character's head moves downward to the lower-left, our model delivers a more precise prediction of the vanishing region near the character's chest.
\begin{table}[t]
\centering
\caption{\textbf{Qualitative result of two Animation Application depending optical flow.} (a) Colorization using BasicPBC~\cite{pbcconference}, (b) In-betweening (interpolation) using AnimeInterp~\cite{animeinterp}.}

\begin{subtable}{0.46\textwidth}
    \centering
    \caption{Testing results of References-based linedrawing colorization on PBC testing set. We substitute RAFT weight of BasicPBC~\cite{pbcconference} and evaluate the colorization performance. Acc: segment-wise accuray. B-MIoU: segment-wise background MIoU.}
    \label{tab:sub_colorization}

    \begin{tabular}{lcccc}
    \hline
    Metrics & Ori  & T+S & T+R & T+LT \\
    \hline
    Acc$~\uparrow$     & 81.28 & 81.44 & 82.65 & 82.94 \\
    BMIoU$~\uparrow$     & 59.38 & 60.72 & 58.72 & 60.82 \\
    \hline
    \end{tabular}
\end{subtable}
\hfill
\begin{subtable}{0.46\textwidth}
    \centering
    \caption{Testing results of Interpolation on anime industry dataset, Cel. We substitute RAFT module of AnimeInterp~\cite{animeinterp} and evaluate the Interpolation performance. PSNR: Peak Signal-to-Noise Ratio. SSIM~\cite{ssim}: Structural Similarity Index.}
    \label{tab:sub_interpolation}
    
    \begin{tabular}{lcccc}
    \hline
    Metrics & Ori & T+S & T+R & T+LT \\
    \hline
    PSNR$~\uparrow$   & 19.16 & 18.22 & 18.93 & 19.63 \\
    SSIM$~\uparrow$   & 0.882 & 0.859 & 0.882 & 0.884 \\
    \hline
    \end{tabular}
\end{subtable}

\end{table}

\subsection{Real Animation Applications.}
\begin{figure}[ht]
\centering
    \begin{subfigure}[b]{\linewidth}
        \centerline{\includegraphics[width=\linewidth]{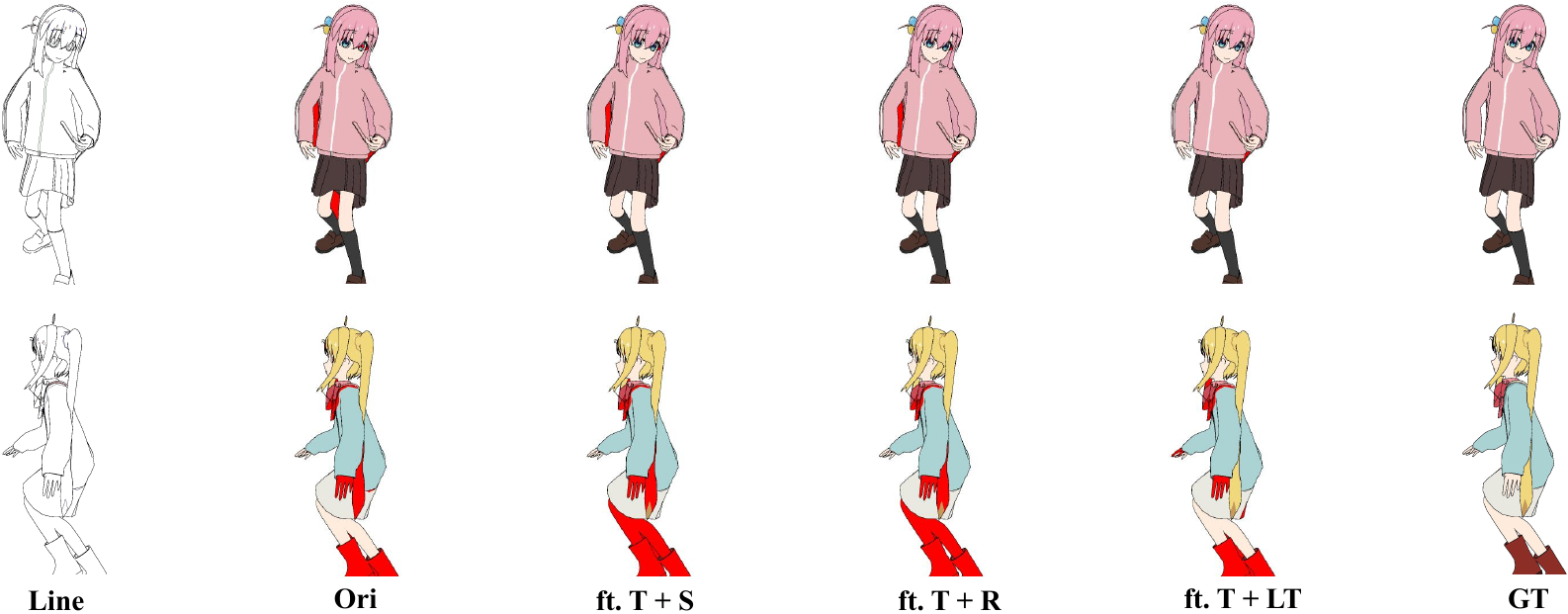}}
        \caption{\textbf{Visualization of line drawing colorization.} The red highlighted areas indicated the incorrect colorization results.}
        \label{fig:figure_interp}
    \end{subfigure}
    \hfill
    \begin{subfigure}[b]{\linewidth}
        \centerline{\includegraphics[width=\linewidth]{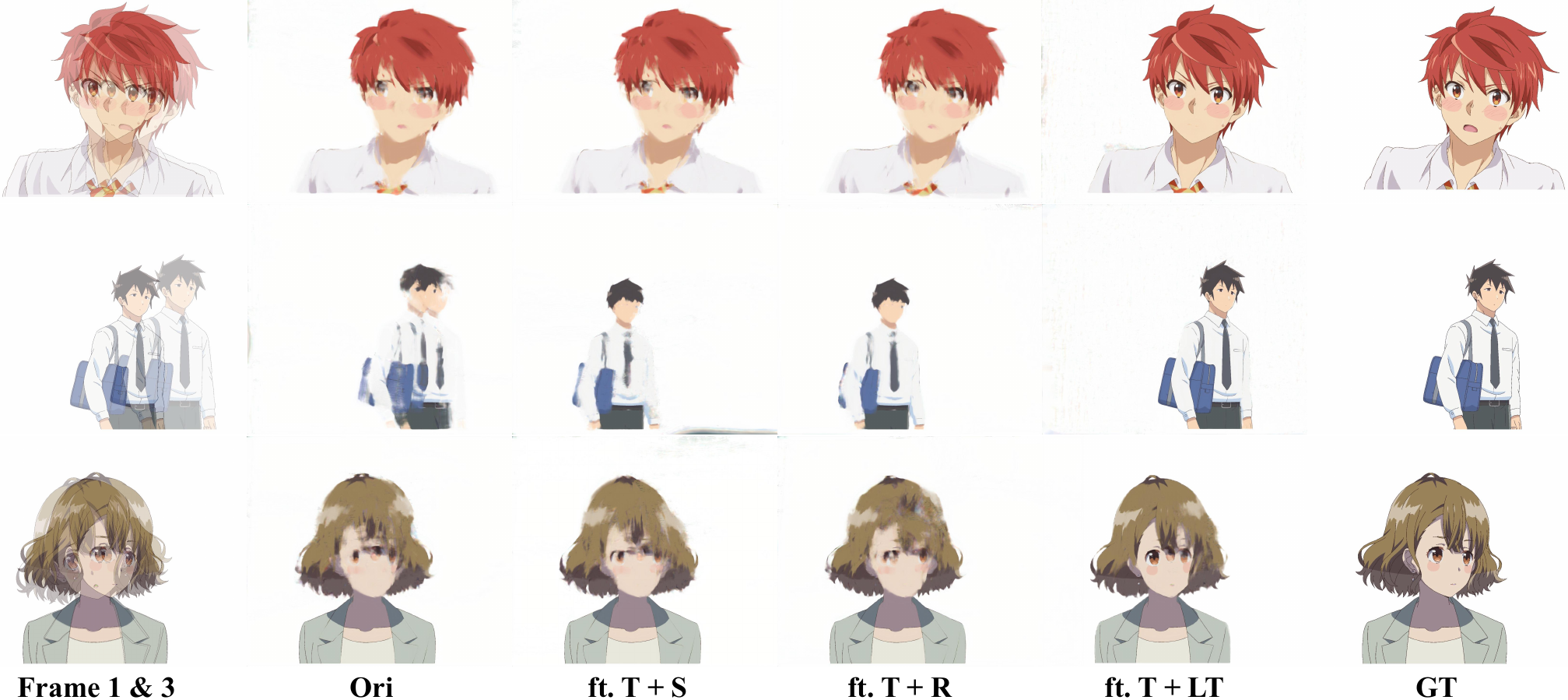}}
        \caption{\textbf{Visualization of frame interpolation on Cel dataset.} Given frames 1~\& 3, the intermediate frame is interpolated using AnimeInterp~\cite{animeinterp} with RAFT finetuned(ft.) on different datasets.}
        \label{fig:figure_pbc}
    \end{subfigure}
\caption{Visualization of two anime applications: (a) Anime Interpolation (b) Anime Interpolation}
\label{fig:figure2}
\end{figure}

To examine the benefits of our dataset for cel anime-industry applications, we chose two animation production methods incorporating optical flow as a component. The first is BasicPBC~\cite{pbcconference}, which employs motion prediction to colorize line drawings using reference frames. The second is AnimeInterp~\cite{animeinterp}, using bidirectional optical flow for interpolating intermediate frames. Two methods align with the Colorization and In-between stages in cel anime production pipelines, respectively. 

\textbf{Line drawing colorization.}
To explore the effectiveness of our dataset for line drawing colorization, we fine-tuned RAFT on different datasets and applied the resulting weights to line-drawing-based colorization. Specifically, we integrated the fine-tuned RAFT models into the optical flow module of BasicPBC~\cite{pbcconference} for evaluation. We measured colorization performance using segment-wise accuracy (Acc) and background mean Intersection over Union (B-MIoU) following~\cite{pbcconference,pbcarticle}. As shown in Tab.~\ref{tab:sub_colorization}, the original RAFT model achieves 81.28 Acc (\%) and 59.38 B-MIoU (\%), which we consider as baseline. Fine-tuning on Sintel~\cite{sintel} slightly improves Acc by 0.16 and increases B-MIoU by 1.34 points. Fine-tuning on AnimeRun~\cite{animerun} leads to a 1.37 increase in Acc, but decreases B-MIoU by 0.66. In contrast, fine-tuning on our LinkTo-Anime dataset significantly boosts both metrics, increasing Acc by 1.66 and B-MIoU by 1.44, demonstrating clearly superior performance.

\textbf{Anime In-between.}
We study the effects of our dataset on the use of anime between. Using RAFT~\cite{raft} models refined with various datasets, we substituted the optical flow module in AnimeInterp~\cite{animeinterp} and tested its efficacy in the Cel dataset. Initially, we converted the Cel data set into a triplet format, similar to ATD-12K~\cite{animeinterp}, and then produced intermediate frames from 1st and 3th frames using each RAFT variant. As per AnimeRun~\cite{animerun}, the interpolation quality is assessed via average PSNR and SSIM compared to the ground truth, as indicated in Tab.~\ref{tab:sub_interpolation}. The model fine-tuned with LT + T demonstrates an improvement in both the PSNR and SSIM metrics compared to the baseline RAFT model, which has a PSNR of 19.16 and an SSIM of 0.882. This contrasts to the performance of other fine-tuned models, which exhibit varying degrees of reduced efficacy across these metrics.

\section{Conclusion}
This paper introduces LinkTo-Anime, a pioneering high-quality optical flow dataset crafted specifically for the anime industry, focusing on cel animation production. Our dataset encompasses both colored frames and line drawings alongside their respective optical flow annotations. This dataset is generated by applying continuous and nuanced movements to meticulously refined open-source 3D models, transforming them into 2D. Beside that, we set up a benchmark using our dataset to assess the effectiveness of various models on both our data and existing data from the animation industry. Our dataset have the potential to support the research of the anime industry automation, improving the efficiency of animations production.

\clearpage
{\small
\bibliographystyle{ieeenat_fullname}
\bibliography{arxiv}
}

\appendix
\newpage
\end{document}